\documentclass[accepted]{uai2026} 
                        
\usepackage[american]{babel}

\usepackage{natbib} 
\usepackage{mathtools} 
\usepackage{booktabs} 
\usepackage{tikz} 

\usepackage{adjustbox}
\usepackage{microtype}
\usepackage{graphicx}
\usepackage{subcaption}
\usepackage{booktabs}
\usepackage{hyperref}
\usepackage{amsmath}
\usepackage{amssymb}
\usepackage{mathtools}
\usepackage{amsthm}
\usepackage{xspace}
\usepackage{subcaption}

\newcommand{\kron}{\otimes}

\renewcommand{\vec}{{\rm vec}}

\renewcommand{\d}{{\rm d}}  

\newcommand{\s}{{\bf s}}
\renewcommand{\t}{{\bf t}}

\renewcommand{\v}{{\bf v}}

\newcommand{\x}{{\bf x}}
\newcommand{\y}{{\bf y}}

\newcommand{\C}{{\bf C}}

\newcommand{\Dcal}{\mathcal{D}}
\newcommand{\Ocal}{\mathcal{O}}
\newcommand{\E}{{\bf E}}

\newcommand{\Gcal}{{\mathcal{G}}}

\newcommand{\I}{{\bf I}}

\newcommand{\K}{{\bf K}}

\newcommand{\Mcal}{{\mathcal{M}}}

\newcommand{\Qcal}{{\mathcal{Q}}}

\newcommand{\N}{\mathcal{N}}  
\newcommand{\gp}{\mathcal{GP}}  

\newcommand{\Q}{{\bf Q}}

\newcommand{\Scal}{{\mathcal{S}}}

\newcommand{\Hcal}{{\mathcal{H}}}
\newcommand{\Fcal}{{\mathcal{F}}}

\newcommand{\boldeta}{\boldsymbol{\eta}}

\newcommand{\bgamma}{\boldsymbol{\gamma}}

\newcommand{\0}{{\bf 0}}

\newcommand{\ben}{\begin{enumerate}}
\newcommand{\een}{\end{enumerate}}

\newcommand{\EE}{\mathbb{E}}

\newcommand{\cmt}[1]{}

\theoremstyle{plain}
\newtheorem{theorem}{Theorem}[section]
\newtheorem{proposition}[theorem]{Proposition}

\theoremstyle{definition}

\theoremstyle{remark}

\newcommand{\ours}{KSTPP\xspace}
\title{Kronecker-Structured Nonparametric Spatiotemporal Point Processes}

\author[1]{Zhitong Xu}
\author[1]{Qiwei Yuan}
\author[1]{Yinghao Chen}
\author[2]{Yan Sun}
\author[3]{Bin Shen}
\author[1]{\href{mailto:<zhe@cs.utah.edu>?Subject=KSTPP}{Shandian Zhe}{}}
\affil[1]{%
    Kahlert School of Computing\\
    The University of Utah
}
\affil[2]{%
    College of Arts \& Sciences\\
    Utah State University
}
\affil[3]{%
    Celonis AI
} 
  
\begin{document}
\maketitle
\begin{abstract}
Events in spatiotemporal domains arise in numerous real-world applications, where uncovering event relationships and enabling accurate prediction are central challenges. Classical Poisson and Hawkes processes rely on restrictive parametric assumptions that limit their ability to capture complex interaction patterns, while recent neural point process models increase representational capacity but integrate event information in a black-box manner, hindering interpretable relationship discovery. To address these limitations, we propose a Kronecker-Structured Nonparametric Spatiotemporal Point Process (\ours) that enables transparent event-wise relationship discovery while retaining high modeling flexibility. We model the background intensity with a spatial Gaussian process (GP) and the influence kernel as a spatiotemporal GP, allowing rich interaction patterns including excitation, inhibition, neutrality, and time-varying effects. To enable scalable training and prediction, we adopt separable product kernels and represent the GPs on structured grids, inducing Kronecker-structured covariance matrices. Exploiting Kronecker algebra substantially reduces computational cost and allows the model to scale to large event collections. In addition, we develop a tensor-product Gauss-Legendre quadrature scheme to efficiently evaluate intractable likelihood integrals. Extensive experiments demonstrate the effectiveness of our framework. The code is released at \url{https://github.com/BayesianAIGroup/KSTPP}.

\end{abstract}
\section{Introduction}

Spatiotemporal events arise in many real-world domains, including weather dynamics, traffic accidents, natural disasters, epidemics, and population migration. Modeling such events for relationship discovery and predictive analysis is crucial. Understanding interactions among events facilitates uncovering the underlying mechanisms driving these phenomena, while accurate prediction enables risk monitoring, early warning, and timely intervention.

Existing spatiotemporal point process models --- though widely used --- face several limitations. Classical Poisson processes assume event independence and thus ignore mutual influences. Hawkes processes~\citep{hawkes1971spectra} introduce self-excitation via triggering effects from past events but typically rely on parametric kernels (e.g., exponential forms), which restrict their ability to capture diverse temporal patterns and inhibitory interactions.

At the other extreme, recent neural point process models directly parameterize the conditional intensity using deep architectures. For example, Neural Hawkes processes~\citep{mei2017neural} and Recurrent Marked Point Processes~\citep{du2016recurrent} encode event histories via recurrent neural networks; Neural Spatial Temporal Point Processes~\citep{chen2021neuralstpp} and Neural Jump Stochastic Differential Equations~\citep{jia2019neural} incorporate continuous latent dynamics; and Transformer Hawkes Processes~\citep{zuo2020transformer} and Self-Attentive Hawkes Processes~\citep{zhang2020self} leverage transformer-based encoders. Although these approaches substantially enhance representational capacity, event interactions are encoded implicitly within latent states, hindering explicit and interpretable relationship discovery.

To address these limitations, we propose \ours, a Kronecker-Structured Nonparametric Spatiotemporal Point Process. Our framework enables transparent and explicit discovery of event-wise relationships while retaining high modeling flexibility to capture complex interaction patterns, including excitation, inhibition, neutrality, and time-varying effects. Our main contributions are summarized as follows:

\begin{itemize}
\item \textbf{Model.} We model the background intensity with a spatial Gaussian process (GP) and the influence kernel as a spatiotemporal GP. The conditional intensity is defined as the superposition of the background intensity and the aggregated influence from past events, followed by a positive link function. This formulation flexibly captures spontaneous occurrences and heterogeneous interaction patterns while maintaining explicit and interpretable representations of event relationships.

\item \textbf{Algorithm.} To enable efficient maximum likelihood training and predictive inference, we employ separable product kernels and represent each GP on structured grids using inducing points, inducing Kronecker-structured covariance matrices. By exploiting Kronecker algebra, covariance operations decompose across input dimensions, substantially reducing computational complexity and enabling scalability to large event collections. To address the intractable integrals arising in likelihood evaluation and predictive density computation, we further develop a tensor-product Gauss-Legendre quadrature scheme. Leveraging the same Kronecker structure, we efficiently evaluate the GPs on structured quadrature grids, enabling tractable numerical evaluation of the required intensity integrals.

\item \textbf{Experiments.} Experiments on three real-world benchmark datasets show that our method consistently outperforms state-of-the-art neural point process models in next-event prediction and achieves competitive performance relative to diffusion-based generative approaches.
Synthetic experiments demonstrate accurate recovery of the underlying intensity functions and interaction patterns. Furthermore, analysis on a real-world earthquake dataset reveals meaningful and interpretable influence structures. Extensive ablation studies demonstrate the effectiveness and robustness of our approach with respect to its key design choices.
\end{itemize}

\section{Preliminaries}

Spatiotemporal point processes naturally extend temporal point processes~\citep{daley2008introduction}. A spatiotemporal point process models random events occurring over a temporal domain $[0,T]$ and a spatial domain $\mathcal{S} \subset \mathbb{R}^2$. An observed event sequence is represented as $\Gamma = \{(t_n, \s_n)\}_{n=1}^N$, where $0 < t_1 < \dots < t_N \le T$ denote the event times and each $\mathbf{s}_n = (x_n, y_n) \in \mathcal{S}$ denotes the spatial location of the $n$-th event.

In this work, we assume a rectangular spatial domain,
\begin{align}
\Scal = [a_x, b_x] \times [a_y, b_y],
\end{align}
although the framework naturally extends to three-dimensional spatial domains when needed. 
The process is characterized by its conditional intensity function 
$\lambda(t, \mathbf{s} \mid \Hcal_t)$, 
defined through the infinitesimal event probability, 
\begin{align}
&\mathbb{P}\big(
\text{an event occurs in } 
[t, t+dt) \times \d\s
\mid \Hcal_t
\big) \notag \\
&=
\lambda(t, \s \mid \Hcal_t)\, \d t \, \d\s, \notag 
\end{align}
where $\d\s$ denotes an infinitesimal spatial area element and $\Hcal_t$ represents the history of events prior to time $t$, i.e., 
\begin{align}
\Hcal_t = \{(t_n, \s_n)|t_n < t\}.
\end{align}
Intuitively, $\lambda(t,\s  \mid \Hcal_t)$ corresponds to the instantaneous event rate at time $t$ and location $\s$ given the past history.
Under standard regularity conditions, the log-likelihood of observing $\Gamma$ is given by
\[
\sum\nolimits_{n=1}^{N} 
\log \lambda(t_n, \s_n  \mid \Hcal_{t_{n}})
-
\int_{0}^{T} 
\int_{\Scal} 
\lambda(t, \s  \mid \Hcal_t) 
\, \d\s \, \d t.
\]


\section{Methodology}

\subsection{Model}
To enable explicit event-wise relationship discovery while retaining high flexibility to capture complex interaction patterns, we model the conditional intensity as 
\begin{align}
    &\lambda(t, x, y\mid \Hcal_t)  \\
    &= \sigma\left(g(x, y) + \sum_{t_n<t} f(t-t_n, x- x_n, y - y_n)\right) \notag
\end{align}
where $g(x, y)$ denotes a latent spatial baseline function capturing spontaneous event occurrences across locations, and $f(\Delta t, \Delta x, \Delta y)$ denotes the influence kernel that characterizes how each past event affects the intensity at $(t, x, y)$. 

Unlike classical Hawkes processes, we do not restrict the influence kernel to be nonnegative. Our formulation allows $f$ to take arbitrary values: $f > 0$ corresponds to excitation, $f < 0$ corresponds to inhibition, and $f \approx 0$ indicates negligible (neutral) influence. 
Because $f$ is defined over temporal and spatial differences, it flexibly models how interaction strength evolves across both time and space.

To ensure positivity of the conditional intensity, we apply a positive link function $\sigma(\cdot)$ to  the superposition of the baseline and influence terms. In this work, we use the SoftPlus function,  $\sigma(z) = \frac{1}{\beta} \log\!\left(1 + e^{\beta z}\right)$, 

where $\beta > 0$ controls the sharpness of the transformation. 

To flexibly estimate $g$ and $f$, we place Gaussian process (GP) priors~\citep{williams2006gaussian} on both functions:
\begin{align}
    g(x, y) &\sim \gp\left(0, \rho(\cdot, \cdot)\right), \notag \\
    f(\Delta t, \Delta x, \Delta y) &\sim \gp\left(0, \kappa(\cdot, \cdot)\right), 
\end{align}
where $\rho$ and $\kappa$ are covariance (kernel) functions for $g$ and $f$, respectively. 

\subsection{Algorithm}\label{sect:algo}
Due to the GP priors over $g$ and $f$, the joint distribution of their values evaluated at observed event times and locations, and at all auxiliary points required for likelihood evaluation (e.g., those for the integral terms) follows a multivariate Gaussian distribution with large covariance matrices defined by $\rho$ and $\kappa$. 
Direct training and inference are therefore computationally prohibitive, requiring $\Ocal(\overline{N}^3)$ time and $\Ocal(\overline{N}^2)$ memory complexity, where $\overline{N}$ denotes the total number of function evaluations involved. The computational cost prevents the model from handling large-scale event datasets.

\paragraph{Kronecker-structured inducing representation.} To overcome this challenge, we construct separable product kernels for both $g$ and $f$:
\begin{align}
    &\rho\left((x, y), (x', y')\right) = \rho_1(x, x')\cdot \rho_2(y, y'), \notag \\
    &\kappa\left((\Delta_t, \Delta_x, \Delta_y), (\Delta_t', \Delta_x', \Delta_y')\right) =  \notag \\
    &\kappa_{0}(\Delta_t, \Delta_t')\cdot \kappa_{1}(\Delta_x, \Delta_x')\cdot \kappa_{2}(\Delta_y,\Delta_y'). \label{eq:prod-kernel}
\end{align}
The widely used squared exponential (SE) kernel is already separable. More generally, different kernels can be chosen along each dimension and multiplied together. This construction corresponds to performing high-dimensional feature mappings along each dimension and taking their tensor product in the latent feature space.

Next, we introduce a structured grid of inducing points. Taking $f$ as an example,  we define the mesh $\Mcal_f = \bgamma_0 \times \bgamma_1 \times \bgamma_2 \subset [0, T] \times [a_x, b_x] \times [a_y, b_y]$, where each $\bgamma_k =\{z^k_1, \ldots, z^k_{m_k}\}$ contains $m_k$ points along dimension $k$. Let $\Fcal$ denote the tensor of function values of $f$ evaluated on $\Mcal_f$. Since $f\sim\gp$, $\vec(\Fcal)$ follows a multivariate Gaussian prior. Under the separable kernel construction~\eqref{eq:prod-kernel}, the covariance matrix admits a Kronecker product structure:
\begin{align}
    p(\Fcal) = \N(\vec(\Fcal) \mid \0, \K_0 \kron \K_1 \kron \K_2),
\end{align}
where $\K_i = \kappa_i(\bgamma_i, \bgamma_i)$ for $i=0,1,2$. Exploiting Kronecker algebra~\citep{kolda2006multilinear}, the log prior decomposes as: 
\begin{align}
    &\log p(\Fcal) =-\frac{1}{2}\sum\nolimits_{i=0}^{2}\frac{m}{m_i}\log|\K_{i}|\notag \\
    &-\frac{1}{2}\vec(\Fcal)^{\top}\vec(\Fcal\times_0\K_{0}^{-1}\times_1\K_{1}^{-1}\times_2\K_{2}^{-1}), \label{eq:log-prior}
\end{align}
where $m= \prod_{k=0}^2 m_k$ is the total number of mesh points, and $\times_i$ denotes the tensor-matrix multiplication at mode $i$. 

Given $\Fcal$, we evaluate $f$ at arbitrary inputs via GP conditional mean (interpolation):
\begin{align}
    &f(\Delta t, \Delta x, \Delta y) = \kappa_0(\Delta t, \bgamma_0)\kappa_1(\Delta x, \bgamma_1) \kappa_2(\Delta y, \bgamma_2) \notag \\
    &\cdot \left(\K_0 \kron \K_1 \kron \K_2\right)^{-1}  \vec(\Fcal)\notag \\
    &= \Fcal \times_0 \boldeta_0  \times_1 \boldeta_1  \times_2 \boldeta_2, \label{eq:gp-int}
\end{align}
where $\boldeta_0 = \kappa_0(\Delta t, \bgamma_0) \K_0^{-1}$, $\boldeta_1 = \kappa_1(\Delta x, \bgamma_1)\K_1^{-1}$ and $\boldeta_2 = \kappa_2(\Delta y, \bgamma_2)\K_2^{-1}$.  Importantly, neither~\eqref{eq:log-prior} nor~\eqref{eq:gp-int} requires explicit  computation of the full $m\times m$ covariance matrix.  All computations are confined to the per-dimensional kernel matrices $\{\K_i\}$, reducing the time complexity from $\Ocal(\prod_{k} m_k^3)$ to $\Ocal(\sum_k m_k^3)$, and the memory complexity from $\Ocal(\prod_k m_k^2)$ to $\Ocal(\sum_k m_k^2)$. This  enables our method to scale to large event datasets encountered in practice. 
  The same construction applies to $g$, using a spatial mesh $\Mcal_g=\v_1\times\v_2$ with corresponding tensor values $\Gcal$.

\paragraph{Training objective.} Given observed event sequences $\Dcal$, the training maximizes the joint log probability:
\begin{align}
    &\log p(\Fcal, \Gcal, \Dcal) = \log p(\Fcal) + \log p(\Gcal) \notag \\
    &+\sum\nolimits_{\Gamma \in \Dcal} \log p(\Gamma | \Fcal, \Gcal). \label{eq:log-joint}
\end{align}
For each sequence $\Gamma = \{(t_n, x_n, y_n)\}_{n=1}^N \in \Dcal$, the log likelihood is 
\begin{align}
    &\log p(\Gamma | \Fcal, \Gcal) =\sum\nolimits_{n=1}^N \log \lambda(t_n, x_n, y_n|\Hcal_{t_n}) \notag \\
    &- \sum_{n=0}^N\int_{t_n}^{t_{n+1}} \int_{a_x}^{b_x}\int_{a_y}^{b_y} \lambda(t, x, y|\Hcal_{t_{n+1}}) \d t \d x \d y, \label{eq:event-ll}
\end{align}
where $t_0 = 0$ and $t_{N+1}=T$. Note that the temporal integral must be evaluated piecewise over each interval $(t_n, t_{n+1})$, since the event history updates immediately after each observed event, causing a discontinuous jump in the conditional intensity $\lambda$.

\paragraph{Tensor-product Gauss-Legendre quadrature.}
The integral terms in the likelihood~\eqref{eq:event-ll} are intractable and do not admit closed-form solutions. While crude Monte Carlo approximation may suffice for stochastic training, reliable prediction of future events (see Section~\ref{sect:pred}) requires high-accuracy evaluation of these integrals. Obtaining such accuracy with Monte Carlo would require a prohibitively large number of samples due to its inherent variance. To address this challenge, we propose a tensor-product Gauss-Legendre quadrature scheme for efficient and high-order numerical integration. 

Specifically, for one-dimensional quadrature rules, the nodes and weights depend only on the integration interval and the chosen rule, and are independent of the integrand. This property allows us to compute the triple integral dimension by dimension using a tensor-product construction. In particular,

\begin{align}
    &\int_{t_n}^{t_{n+1}} \lambda(t, x, y|\Hcal_{t_{n+1}}) \d t \d x \d y \notag \\
    &\approx \sum_{i} w^0_i \int_{a_x}^{b_x} \int_{a_y}^{b_y}\lambda(\hat{t}_i, x, y|\Hcal_{t_{n+1}})  \d x \d y \notag \\
    &\approx \sum_{i} w^0_i \sum_j w^1_j \int_{a_y}^{b_y} \lambda(\hat{t}_i, \hat{x}_j, y|\Hcal_{t_{n+1}}) \d y  \notag \\
    &\approx \sum_{i} w^0_i \sum_j w^1_j  \sum_k w^2_k \lambda(\hat{t}_i, \hat{x}_j, \hat{y}_k|\Hcal_{t_{n+1}}) \notag \\
    & = \sum_{i,j,k} w_i^0 w_j^1 w_k^2 \cdot \lambda(\hat{t}_i, \hat{x}_j, \hat{y}_k|\Hcal_{t_{n+1}}), \label{eq:tensor-prod-quad}
\end{align}
where $\{w_i^0, \hat{t}_i\}$, $\{w_j^1, \hat{x}_j\}$, and $\{w_k^2, \hat{y}_k\}$ denote the quadrature nodes and weights along the temporal and spatial dimensions, respectively. Because the nodes and weights are separable across dimensions and independent of $\lambda$, the multi-dimensional integral reduces to evaluating $\lambda$ on the structured quadrature grid, 
\[
\Qcal = \hat{\t} \times \hat{\x} \times \hat{\y}, \; \hat{\t}=\{\hat{t}_i\}, \; \hat{\x} = \{\hat{x}_j\}, \; \hat{\y} = \{\hat{y}_k\},
\]
followed by a tensor-weighted inner product with the separable weight tensor $\{w_i^0 w_j^1 w_k^2\}_{i,j,k}$.

Evaluating $\lambda$ on the quadrature grid $\Qcal$ requires computing the baseline component $g$ and the influence kernel $f$ at all grid locations. This can be carried out efficiently by leveraging the Kronecker structure again. In particular, $f(\Qcal) = \Fcal \times_0 \kappa_0(\hat{\t}, \bgamma_0) \K_0^{-1} \times_1 \kappa_1(\hat{\x}, \bgamma_1) \K_1^{-1} \times_2 \kappa_2(\hat{\y}, \bgamma_2) \K_2^{-1}$, and the evaluation of $g$ on $\Qcal$ follows analogously using its spatial mesh.

We adopt Gauss-Legendre rules along each dimension due to their high accuracy for smooth integrands, typically requiring only a small number of nodes. For a general interval $[a, b]$, the quadrature nodes and weights are obtained via a linear transformation of the standard rule on $[-1, 1]$: $\hat{z}_k = \frac{b-a}{2} \xi_k + \frac{b+a}{2}, w_k = \frac{b-a}{2} \alpha_k$, where  $\{\xi_k\}$  and $\{\alpha_k\}$ denote the standard Gauss-Legendre nodes and weights on $[-1, 1]$. 

Combining the Kronecker-structured GP representation with the tensor-product quadrature scheme allows efficient computation of both the log prior and the log likelihood for each event sequence in~\eqref{eq:log-joint}. Training is performed using stochastic mini-batch optimization over randomly sampled sequences to estimate $\Fcal$, $\Gcal$, and kernel parameters. The full model pipeline is summarized in Appendix Figure~\ref{fig:pipeline} and Section~\ref{app:pipeline}.

\noindent\textbf{Computational Complexity.}
The time complexity for processing a mini-batch of $B$ sequences, each of length $N$, is $$\Ocal\left(\sum_{k=0}^2 m_k^3 + \sum_{k=1}^2 v_k^3 + BN\left(m(\sum_{i=0}^2 q_i) + v(\sum_{i=1}^2 q_i)\right)\right),$$ where $m=\prod_{k=0}^2m_k$ denotes the total number of mesh points in $\Mcal_f$,  $v=\prod_{k=1}^2v_{k}$ denotes the total number of  mesh points in $\Mcal_g$, and $q_i$ is the number of quadrature nodes along each dimension $i$. The memory complexity is $$\Ocal(\sum\nolimits_{k=0}^2 m_k^2 +\sum\nolimits_{k=1}^2 v_k^2 + m + v + BNq),$$ where $q=\prod_i q_i$ is the total number of nodes in the tensor-product grid $\Qcal$. The dominant memory cost arises from storing the per-dimension kernel matrices and maintaining function evaluations at mesh points, quadrature nodes, and observed events. Overall, the cubic terms depend only on per-dimension mesh sizes rather than the total number of events, enabling scalable mini-batch training and inference.

\subsection{Prediction}\label{sect:pred}
Given a sequence of $N$ observed events $\Hcal = \{(t_1, x_1, y_1), \ldots, (t_N, x_N, y_N)\}$, we aim to predict the time and location of the next event $(t_{N+1}, x_{N+1}, y_{N+1})$.

\paragraph{Predicting the next event time.} The conditional density of the next arrival time is given by the standard point process formulation:
\begin{align}
    p(t_{N+1} \mid \Hcal) = \lambda(t_{N+1}|\Hcal) \exp\left(-\int_{t_N}^{t_{N+1}}\lambda(t|\Hcal) \d t\right), \notag 
\end{align}
where the temporal marginal intensity is 
\begin{align}
\lambda(t|\Hcal) = \int_{a_x}^{b_x}\int_{a_y}^{b_y} \lambda(t, x, y|\Hcal) \d x \d y. \label{eq:lambda_t}
\end{align}
We use the posterior mean as the point prediction of $t_{N+1}$. Let $\tau = t_{N+1} - t_N > 0$. Then 
\begin{align}
    \EE[t_{N+1}|\Hcal] = t_N + \EE[\tau|\Hcal].
\end{align}
Using integration by parts, the conditional expectation of $\tau$ can be written as 
\begin{align}
    \EE[\tau|\Hcal] &= \int_0^\infty e^{-\Lambda(\tau)} \d \tau, \label{eq:outer} \\
    \Lambda(\tau) &= \int_0^\tau \lambda(t_N + u|\Hcal) \d u. \label{eq:inner}
\end{align}
The evaluation of $\lambda(t \mid \Hcal)$ and the inner integral~\eqref{eq:inner} is performed using the same tensor-product quadrature method described in Section~\ref{sect:algo}. 

To compute the improper integral in~\eqref{eq:outer}, we apply the transformation
\[
u = \frac{\tau}{1+\tau}, \quad \tau = \frac{u}{1-u},
\]
which maps $\tau \in (0, \infty)$ to $u \in (0, 1)$. This yields
\begin{align}
    \EE[\tau|\Hcal] = \int_0^1 \frac{e^{-\Lambda\left(u/(1-u)\right)}} {(1-u)^2} \d u. \label{eq:definite}
\end{align}
We then apply Gauss-Legendre quadrature to evaluate~\eqref{eq:definite}. To further improve time prediction accuracy and mitigate approximation errors introduced by numerical quadrature, we optionally learn a separate MLP head for inter-arrival time prediction, similar in spirit to the approach of~\citet{zuo2020transformer}. Specifically, we evaluate the integrand in~\eqref{eq:definite} at the quadrature nodes and use these values as input to an MLP. A Softplus activation is applied to the network output to ensure positivity. This prediction head is trained separately after the main model has converged. Intuitively, the MLP learns a data-driven correction to the numerical quadrature, compensating for residual approximation errors that are difficult to capture analytically. Empirically, we find that this refinement consistently yields a modest improvement in next-event time prediction over the plain quadrature estimator in~\eqref{eq:definite}.

\paragraph{Predicting the event location.} Given the predicted time $t_{N+1}$, the conditional spatial density is 
\[
p(x, y|t_{N+1}) = \frac{\lambda(t_{N+1}, x, y \mid \Hcal)}{\lambda(t_{N+1} \mid \Hcal)}.
\]
We use the conditional expectations as point predictions:
\begin{align}
    \EE[x|t_{N+1}] &= \int_{a_x}^{b_x}\int_{a_y}^{b_y} x\, p(x, y|t_{N+1}) \d x \d y \notag \\
    \EE[y|t_{N+1}] &= \int_{a_x}^{b_x}\int_{a_y}^{b_y} y\, p(x, y|t_{N+1}) \d x \d y. 
\end{align}
These integrals are evaluated using the same tensor-product quadrature scheme described in Section~\ref{sect:algo}.

\section{Related Work}
\label{sec:related}
A rich body of work has been devoted to temporal point processes. Early developments include Poisson processes~\citep{lloyd2015variational} and their applications in tensor decomposition~\citep{schein2015bayesian,Schein:2016:BPT:3045390.3045686,schein2019poisson}. Hawkes processes (HPs)~\citep{hawkes1971spectra} subsequently gained significant attention due to their ability to capture mutual excitation among events, e.g.,~\citep{blundell2012modelling,du2015dirichlet,wang2017predicting,yang2017decoupling,xu2018benefits,zhe2018stochastic}.
Recent works have extended classical Hawkes processes with more flexible influence kernels capable of modeling inhibitory effects~\citep{pan2020scalable,wang2020self}. \citet{pan2021self} proposed a nonparametric kernel that captures general temporal influence decay patterns beyond standard exponential decay. More recently, \citet{xu2026structured} introduced a product-form neural kernel combining a signed interaction network with a delay-aware monotonic network, enabling flexible modeling of both temporal decay and delayed influences.

In parallel, conditional intensities have been modeled using deep neural architectures. Neural Hawkes Processes (NHP)~\citep{mei2017neural} encode event history via LSTM states~\citep{hochreiter1997long}, with the intensity parameterized as a function of the hidden state. Recurrent Marked Temporal Point Processes (RMTPP)~\citep{du2016recurrent} adopt a similar recurrent architecture while explicitly modeling event marks (types). Transformer-based approaches~\citep{zhang2020self,zuo2020transformer} treat each event as a token and use causal attention mechanisms to aggregate historical information, with the conditional intensity parameterized on top of token representations.

Extending classical Poisson and Hawkes processes to the spatiotemporal setting is conceptually straightforward. Recently,~\citet{chen2021neuralstpp} proposed a neural spatiotemporal point process model that, similar to NHP and RMTPP, models intensity jumps using recurrent neural networks. To enable continuous-time evolution between events, they incorporate neural ordinary differential equations (ODEs)~\citep{chen2018neural}. In contrast,~\citet{jia2019neural} model hidden state dynamics using neural stochastic differential equations (SDEs).
~\citet{pmlr-v168-zhou22a,NEURIPS2023_9d30c2de} extended spatiotemporal Hawkes processes. The model of~\citet{pmlr-v168-zhou22a} preserves the parametric Hawkes structure while estimating the triggering kernel parameters via transformer encodings of historical events. The work of~\citet{NEURIPS2023_9d30c2de} introduces a neural triggering kernel based on monotonic networks~\citep{sill1997monotonic} allowing exact likelihood integration. However, both approaches remain within the Hawkes framework and are therefore limited to modeling excitatory interactions, making them unable to capture inhibitory event dynamics commonly observed in practice. More recently,~\citet{10.1145/3580305.3599511} proposed bypassing intensity modeling entirely by using diffusion-based generative models~\citep{ho2020denoising} to directly generate event sequences. While demonstrating strong predictive performance, such approaches sacrifice the ability to explicitly model and calibrate the conditional intensity, which is critical for applications such as risk monitoring and survival analysis --- central objectives in point process modeling.

The computational advantages of Kronecker product structures have been widely recognized in scalable Gaussian process and kernel methods~\citep{saatcci2012scalable,wilson2015kernel,izmailov2018scalable,zhe2019scalable}. For example,~\citet{xu2012infinite,zhe2016dintucker} exploited Kronecker structure in nonparametric tensor factorization models. More recently, physics-informed machine learning methods~\citep{fang2023solving,xutoward2025} have leveraged Kronecker structure for efficient nonlinear PDE solving. To the best of our knowledge, our work is the first to integrate Kronecker-structured Gaussian process representations into spatiotemporal point process modeling, enabling scalable training and flexible intensity modeling.

\section{Experiments}
\subsection{Synthetic Data}\label{sect:expr-syn}
We first evaluated \ours on synthetic datasets designed to validate its ability to capture both excitation and inhibition effects. We constructed two spatiotemporal point processes, each incorporating excitation and inhibition mechanisms driven by past events. The conditional intensity follows a form resembling a spatiotemporal Hawkes process:
\begin{align}
\lambda(t,x, y|\Hcal_t) = \lambda_0 + \sum_{t_n<t} c_n e^{-\beta(t-t_n)}\frac{1}{2\pi \sigma^2}e^{-\frac{d_n^2}{2\sigma^2}}~\label{eq:simu}  
\end{align}
where $d_n = \sqrt{(x-x_n)^2 + (y-y_n)^2}$ and $\beta, \sigma>0$. In the first process, denoted as \textbf{SYN1}, the influence strength $c_n$ depends on the temporal lag. When $t-t_n<1$, we set $c_n=1.0$ to induce excitation; when $t-t_n\ge 1$, we set $c_n=-2.0$, introducing inhibitory effects. The decay parameter and spatial bandwidth were set to $\beta = 2.0$ and $\sigma=0.3$, respectively. In the second process, denoted as \textbf{SYN2}, the interaction type depends on spatial distance. When $d_n>1.0$, we set $c_n=1.0$, enabling excitation from distant events. When $d_n<1.0$, we set $c_n=-0.3$, modeling local inhibition. The other parameters were set to $\beta=1.5$ and $\sigma=0.5$. For both processes, the time horizon was set to $T = 50$ with base rate $\lambda_0=2$. We generated 2,300 sequences for training, 100 sequences for validation, and another 100 sequences for testing, using Ogata's thinning algorithm~\citep{ogata1981lewis}. 

We compared \ours against several popular and state-of-the-art point process models. 
(1) Spatiotemporal Hawkes Process (STHP): adopts the same conditional intensity form as in~\eqref{eq:simu} but restricts all parameters to be positive (excitation-only). Parameters are optimized in the log domain to enforce positivity.
(2) Neural Spatiotemporal Point Process (NSTPP)~\citep{chen2021neuralstpp}: integrates historical events via RNN states, modeling continuous-time dynamics with neural ODEs and event-triggered updates through GRU-style gating.
(3) Deep Spatiotemporal Point Process (DeepSTPP)~\citep{pmlr-v168-zhou22a}: retains a parametric excitation-only Hawkes intensity while using a transformer encoder to estimate kernel parameters.
We also include two neural temporal point process models:
(4) Neural Hawkes Process (NHP)~\citep{mei2017neural}, which encodes event history using a continuous-time LSTM; and
(5) Transformer Hawkes Process (THP)~\citep{zuo2020transformer}, which applies a transformer to model temporal dependencies.

Our method was implemented in PyTorch and trained using the Adam optimizer with a learning rate of $10^{-3}$. The mini-batch size was set to one. We set $\beta=1$ in the SoftPlus transformation. We used 12 Gauss-Legendre quadrature nodes per dimension and adopted the squared exponential (SE) kernel for the influence function $f$. STHP was trained using Adam with the same learning rate. For the remaining baselines, we used the official open-source implementations and default hyperparameter settings.

\begin{table}[!htp]\centering
\caption{ Relative $L_2$ error of the learned marginal  intensity $\lambda(t|\Hcal_t)$ on synthetic datasets. The smallest error is shown in  bold.}\label{tab:lam-t-l2}

\centering
\begin{tabular}{lcc}
\toprule
&\textbf{SYN1} & \textbf{SYN2} \\\midrule
STHP & 1.45e-01 $\pm_{2.41e-02}$ &  8.42e-02 $\pm_{1.87e-02}$\\
NSTPP & 5.57e-02 $\pm_{6.50e-03}$ & 2.99e-02 $\pm_{4.78e-03}$ \\
DeepSTPP & 1.25e-01 $\pm_{1.89e-02}$ & 7.77e-02 $\pm_{1.52e-02}$ \\
NHP & 1.35e-01 $\pm_{3.08e-02}$  & 2.34e-02 $\pm_{4.37e-03}$ \\
THP & 8.14e-01 $\pm_{7.10e-02}$ & 1.91e-01 $\pm_{2.87e-02}$\\
\ours & \textbf{4.44e-02} $\pm_{3.96e-03}$ & \textbf{2.00e-02} $\pm_{3.50e-03}$\\
\bottomrule
\end{tabular}
\end{table}

\begin{figure}[t]
	\centering
	\includegraphics[width=\columnwidth]{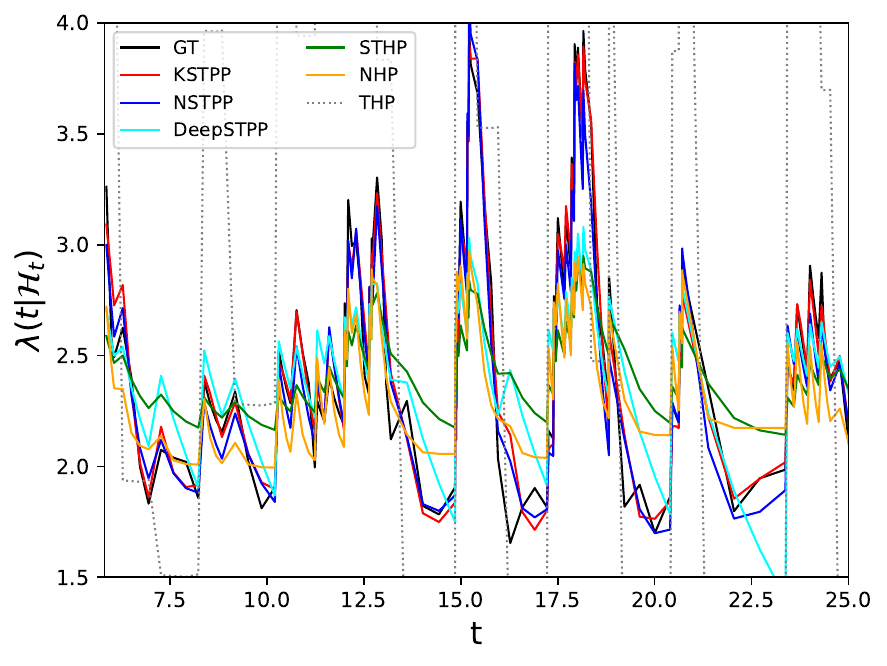}	
	\caption{Temporal conditional intensity on example test sequence from \textbf{SYN1}.}\label{fig:lambda-t-syn1}
\end{figure}

\begin{table}
\centering

\caption{ Relative $L_2$ error  of the learned spatiotemporal intensity $\lambda(t, x,y|\Hcal_t)$ on synthetic datasets.\cmt{The smallest error is shown bold.}}\label{tab:lam-xy}

\begin{tabular}{lcc}\toprule
&\textbf{SYN1} & \textbf{SYN2} \\\midrule
STHP & 1.77e-01 $\pm_{7.44e-02}$ &  1.23e-01$\pm_{4.73e-02}$\\
NSTPP & 8.68e-01 $\pm_{9.41e-03}$ & 8.64e-01 $\pm_{3.26e-03}$ \\
DeepSTPP &3.78e-01 $\pm_{3.65e-02}$ & 3.87e-01 $\pm_{2.47e-02}$ \\
\ours & \textbf{6.52e-02} $\pm_{4.80e-03}$ & \textbf{3.67e-02} $\pm_{1.43e-02}$ \\
\bottomrule
\end{tabular}
\end{table}
\begin{figure*}[!h]
	\centering
	\includegraphics[width=\textwidth]{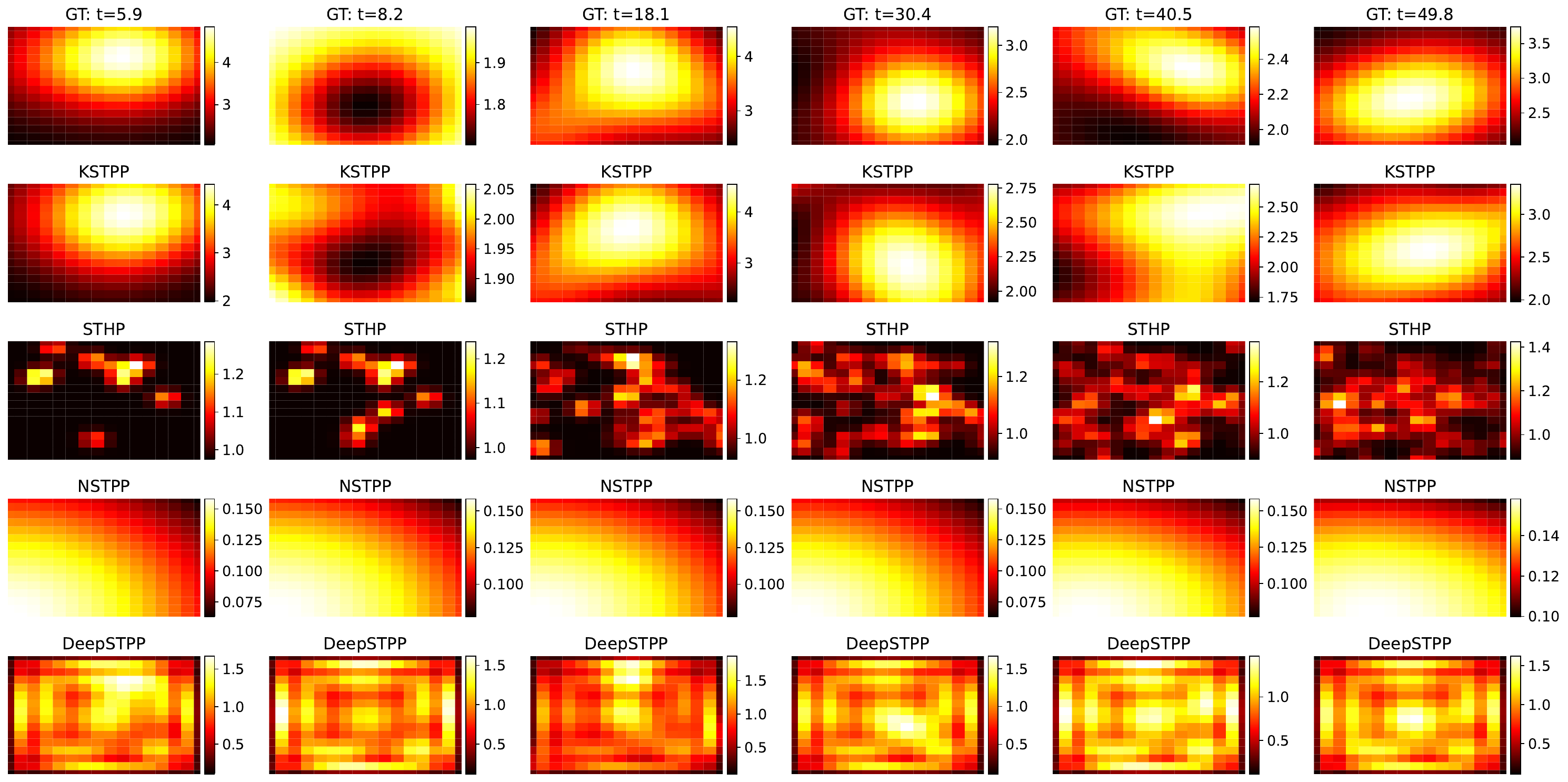}
	\caption{Spatial conditional intensity $p(x,y \mid t,\Hcal_t)=\lambda(t,x,y \mid \Hcal_t)/\lambda(t \mid \Hcal_t)$ on an example test sequence from \textbf{SYN1}. GT denotes the ground-truth.}
	\label{fig:spatial-syn1}
\end{figure*}
\paragraph{Intensity Recovery.}
We first examined whether each method could recover the ground-truth intensity. Since NHP and THP are designed for purely temporal point processes, we compared the marginal conditional intensity $\lambda(t \mid \Hcal_t)$ across all methods.
Specifically, for each test sequence, we evaluated the conditional intensity at every observed event time as well as at three equally spaced time points between successive events. For each sequence, we computed the relative $L_2$ error with respect to the ground-truth intensity, and report the mean and standard deviation across all test sequences. As shown in Table~\ref{tab:lam-t-l2}, \ours consistently achieves the lowest relative $L_2$ error, indicating superior intensity recovery.
Figure~\ref{fig:lambda-t-syn1} and Appendix Figure~\ref{fig:lambda-t-syn2} visualize the recovered temporal intensity for an example test sequence from each synthetic dataset. In \textbf{SYN1}, the intensity curves estimated by STHP and DeepSTPP roughly capture the overall shape of the ground truth, albeit with noticeable inaccuracies. However, in \textbf{SYN2}, their estimates deviate substantially from the true intensity.
This difference may be explained by the strength of inhibition in the two datasets. In \textbf{SYN1}, the inhibition effect is relatively weak --- it appears only when $t>1$ in~\eqref{eq:simu}, and the magnitude of the inhibitory kernel is small. In contrast, \textbf{SYN2} exhibits strong local inhibition within a small spatial range ($d<1$). Since STHP and DeepSTPP rely on fixed kernel forms, model misspecification under strong inhibition can lead to degraded intensity estimates.
NHP and NSTPP demonstrate strong approximation capability, reflecting their expressive modeling capacity. In contrast, THP exhibits substantial deviations from the ground-truth intensity. Our method consistently produces intensity estimates that closely match the ground-truth across both datasets. Moreover, unlike excitation-only models, \ours is capable of identifying both excitation and inhibition effects, as demonstrated later in the influence kernel estimation results.

We next evaluate recovery of the spatial conditional intensity. For each dataset (\textbf{SYN1} and \textbf{SYN2}), we randomly select a test sequence and examine the spatial conditional intensity at several representative time points (Figure~\ref{fig:spatial-syn1} and Appendix Figure~\ref{fig:spatial-syn2}).
As shown, \ours closely matches  both the shape and magnitude of the ground-truth spatial intensity. In contrast, STHP, NSTPP, and DeepSTPP exhibit noticeable deviations and fail to recover key spatial structures. In particular, STHP and DeepSTPP produce multiple spurious local modes, resembling mixture-like densities. This behavior may stem from their use of additive, strictly  positive parametric triggering kernels. Although NSTPP yields smoother spatial intensity estimates, its recovered structure and scale do not align well with the ground truth.

Table~\ref{tab:lam-xy} reports the average relative $L_2$ error of the full spatiotemporal intensity evaluated on a $16 \times 16$ uniform spatial grid. \ours achieves the lowest error across both datasets, quantitatively confirming its improved intensity recovery. Overall, these results highlight the advantage of \ours in recovering both the temporal and spatial components of the underlying intensity.

\begin{figure}[!h]
	\centering
	\includegraphics[width=\columnwidth]{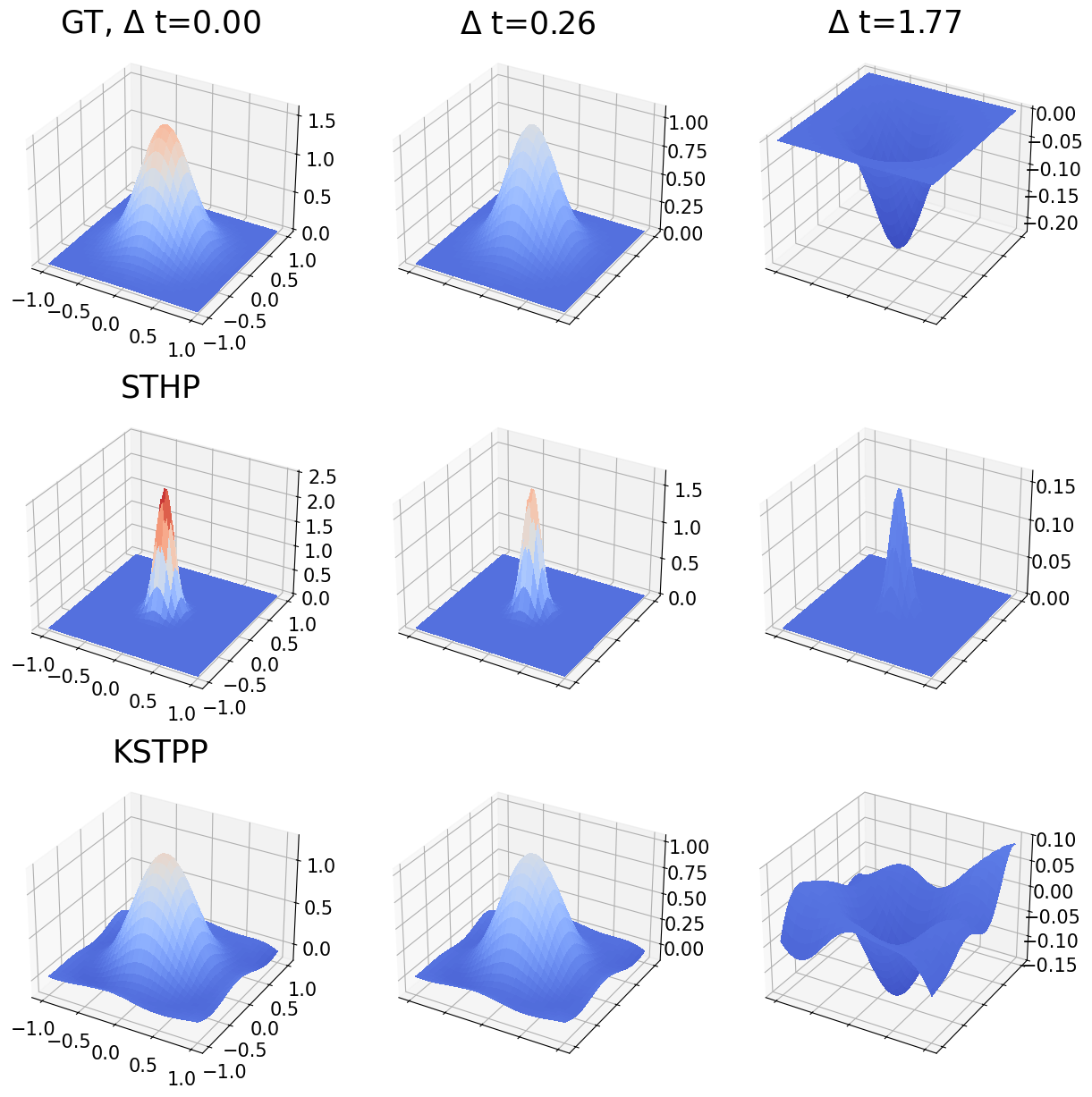}
	\caption{Learned influence kernel from \textbf{SYN1}.}
	\label{fig:influence-kernel-syn1}
\end{figure}
\begin{table*}[th]
\centering
\caption{Root-mean-square error (RMSE) and Euclidean distance for next-event time and location prediction. The best two results are highlighted in bold.}\label{tbl:pred}
\begin{tabular}{c|cc|cc|cc}
\hline
& \multicolumn{2}{c|}{Earthquake} & \multicolumn{2}{c|}{COVID-19} & \multicolumn{2}{c}{Citibike} \\ \cline{2-7} 
Model            & Spatial $\downarrow$   & Temporal $\downarrow$  & Spatial $\downarrow$   & Temporal $\downarrow$  & Spatial $\downarrow$   & Temporal  $\downarrow$ \\ \hline
RMTPP              &   -   &   0.424{\scriptsize $\pm$0.009}     &  -      &     1.32{\scriptsize $\pm$0.024}       &  -       &      2.07{\scriptsize $\pm$0.015}      \\
NHP               &  -   &   1.86{\scriptsize $\pm$0.023}   &   -  &  2.13{\scriptsize $\pm$0.100}    &    -  &  2.36{\scriptsize $\pm$0.056}    \\   
THP              &  -   &   2.44{\scriptsize $\pm$0.021}   &   -  &  0.611{\scriptsize $\pm$0.008}    &    -  &  1.46{\scriptsize $\pm$0.009}\\          \hline

Poisson  & 9.45{\scriptsize $\pm$0.000} & 0.412{\scriptsize $\pm$0.000} & 0.818{\scriptsize $\pm$0.000} & 0.113{\scriptsize $\pm$0.000} & 0.452{\scriptsize $\pm$0.000} & 0.239{\scriptsize $\pm$0.000} \\
STHP & 8.35{\scriptsize $\pm$0.252} & 0.424{\scriptsize $\pm$0.018} & 0.422{\scriptsize $\pm$0.000} & \textbf{0.100}{\scriptsize $\pm$0.001} & 0.032{\scriptsize $\pm$0.000} & 0.633{\scriptsize $\pm$0.126} \\
NJSDE   &    9.98{\scriptsize $\pm$0.024} &   0.465{\scriptsize $\pm$0.009} &  0.641{\scriptsize $\pm$0.009}  &  0.137{\scriptsize $\pm$0.001} &   0.707{\scriptsize $\pm$0.001}    &  0.264{\scriptsize $\pm$0.005} \\ 
NSTPP      &   8.11{\scriptsize $\pm$0.000}        &  0.547{\scriptsize $\pm$0.010}  &   0.560{\scriptsize $\pm$0.000}   &   0.145{\scriptsize $\pm$0.002}   &    0.705{\scriptsize $\pm$0.000}&  0.355{\scriptsize $\pm$0.013}  \\ 
DeepSTPP        & 9.20{\scriptsize $\pm$0.000}   &   \textbf{0.341{\scriptsize $\pm$0.000}}  &  0.687{\scriptsize $\pm$0.000} &  0.197{\scriptsize $\pm$0.000}     &  0.044{\scriptsize $\pm$0.000}    &   0.234{\scriptsize $\pm$0.000} \\
DSTPP   &   \textbf{6.77{\scriptsize $\pm$0.193}}  &  0.375{\scriptsize $\pm$0.001}    & \textbf{0.419{\scriptsize $\pm$0.001}} &        \textbf{0.093{\scriptsize $\pm$0.000}}  & \textbf{0.031{\scriptsize $\pm$0.000}}   &  \textbf{0.200{\scriptsize $\pm$0.002}}\\ \hline 
KSTPP (Ours) & \textbf{6.72{\scriptsize $\pm$0.014}} & \textbf{0.372{\scriptsize $\pm$0.000}} & \textbf{0.392{\scriptsize $\pm$0.000}} & \textbf{0.100{\scriptsize $\pm$0.000}} & \textbf{0.031{\scriptsize $\pm$0.000}} & \textbf{0.206{\scriptsize $\pm$0.000}}\\ \hline 
\end{tabular}
\end{table*}

\textbf{Influence Kernel Estimation.}
We then evaluated the learned influence kernel $f$ produced by \ours. For both \textbf{SYN1} and \textbf{SYN2}, we select three representative time lags and visualize the corresponding learned  kernels. For comparison, we also report the kernel estimated by STHP. As shown in Figure~\ref{fig:influence-kernel-syn1} and Appendix Figure~\ref{fig:influence-kernel-syn2}, \ours effectively recovers both excitation and inhibition patterns. Because our model applies a SoftPlus transformation to enforce non-negativity of the conditional intensity, the learned kernel --- under this nonlinear link --- does not exactly coincide with the ground-truth kernel specified under a purely linear additive formulation. Nevertheless, the key structural characteristics, including the sign, modal location, and spatial spread of the influence, are well preserved. In contrast, although STHP adopts an additive formulation consistent with the data-generating process, it is restricted to excitation-only mechanisms. Consequently, its learned kernel exhibits substantial deviation from the ground truth, particularly in regions where inhibitory effects dominate.

\subsection{Predictive Performance}
Next, we evaluate the predictive performance of \ours on three real-world benchmark datasets: \textbf{Earthquake}~\citep{chen2021neuralstpp}, \textbf{Covid-19}~\citep{chen2021neuralstpp}, and \textbf{Citibike}~\citep{10.1145/3580305.3599511}. These datasets cover seismic activity, epidemic spread, and urban mobility, providing diverse spatiotemporal event dynamics. Detailed descriptions, data splits, and statistics are provided in Appendix~\ref{sect:dataset}.

In addition to the methods introduced in Section~\ref{sect:expr-syn}, we included: (6) Diffusion Spatiotemporal Point Process (DSTPP)~\citep{10.1145/3580305.3599511}, a recent diffusion-based generative model that directly generates event sequences without explicitly modeling the conditional intensity. (7) Neural Jump Stochastic Differential Equations (NJSDE)~\citep{jia2019neural}, which summarizes event history into latent states and models state dynamics via neural SDEs with jumps induced by event arrivals. (8) Homogeneous Poisson Process (PP) as a classical baseline. For \ours, the covariance functions for  $g$ and $f$ were selected from either the SE kernel or Mat\'ern kernel with smoothness parameter $\nu=5/2$. The number of quadrature nodes per dimension was chosen from \{8, 12, 16\}. We further employed a lightweight two-layer MLP prediction head with ReLU or GELU activations for next-event time prediction. Full implementation details and hyperparameter choices are provided in Appendix~\ref{app:impl}.

We adopted the same training, validation, and test splits as provided in~\citep{10.1145/3580305.3599511} for all three datasets. Following~\citep{chen2021neuralstpp,10.1145/3580305.3599511}, each experiment was repeated five times with different random initializations. We evaluated: Root-Mean-Square Error (RMSE) for next-event time prediction, and Euclidean distance for next-event location prediction. We report the mean and standard deviation across runs in Table~\ref{tbl:pred}. For baselines other than PP and STHP, results were directly taken from~\citep{10.1145/3580305.3599511}. Although we conducted additional hyperparameter tuning for these methods, we were unable to outperform the reported results. To ensure fairness, we therefore compare against the optimized results reported in~\citep{10.1145/3580305.3599511}.

As shown in Table~\ref{tbl:pred}, \ours consistently achieves top performance in both event time and location prediction. Its predictive accuracy is comparable to the diffusion-based model DSTPP and often outperforms other neural point process methods by a substantial margin. In particular, \ours achieves the highest location prediction accuracy across all datasets and the second-best performance in time prediction. These results demonstrate that, although \ours adopts a more structured modeling framework aimed at improved event relationship discovery, it maintains strong predictive performance that is competitive with, and in many cases superior to, existing neural point process models.
\begin{figure}
	\centering
	\includegraphics[width=0.8\columnwidth]{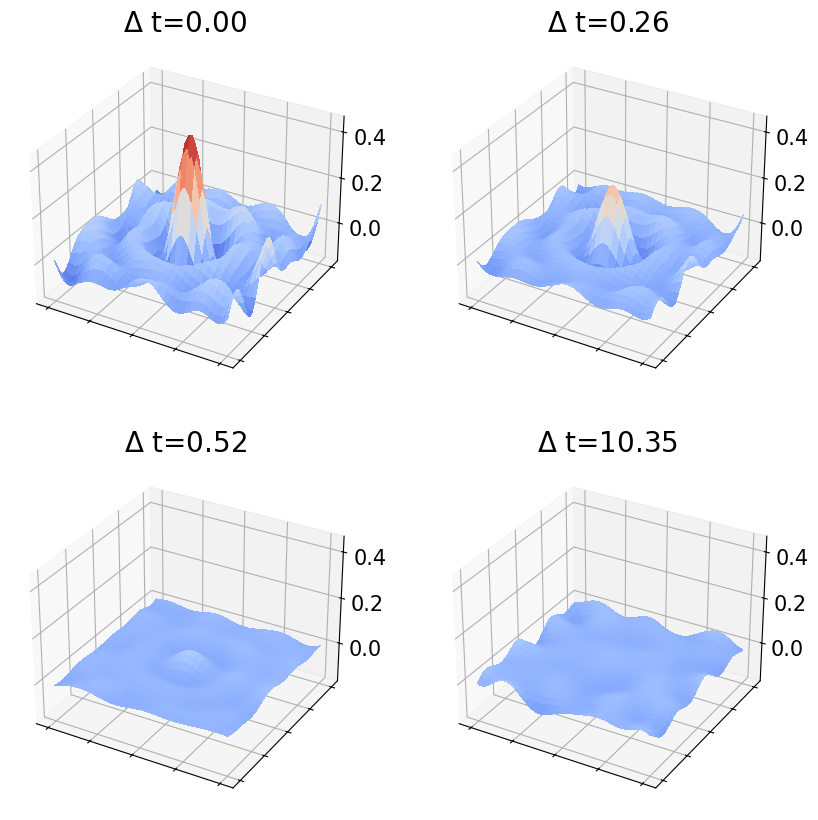}
	\caption{Learned influence kernel by \ours on \textbf{Earthquake}.}
	\label{fig:influence-earth}
\end{figure}

\subsection{Pattern Discovery}

Finally, we investigated the learned influence kernel of \ours on the real-world Earthquake dataset to examine whether the model reveals meaningful spatiotemporal interaction patterns. 

We visualize the learned kernel $f(\Delta t, \Delta x, \Delta y)$ at four representative time lags, $\Delta t \in \{0, 0.26, 0.52, 10.35\}$, across the spatial domain. As shown in Figure~\ref{fig:influence-earth}, when $\Delta t$ is small, the influence exhibits strong excitation in nearby regions (i.e., small $|\Delta x|$ and $|\Delta y|$). As the time lag increases, the overall magnitude of the influence decays. When $\Delta t\ge 0.52$, the kernel values across spatial distances approach zero.
This behavior is consistent with well-established seismic dynamics: earthquakes tend to trigger short-term aftershocks in spatially proximate regions, while such triggering effects decay over time, as described by the Omori-Utsu law and ETAS models~\citep{utsu1995centenary,ogata1988statistical}.
Interestingly, the learned kernel also reveals localized inhibition effects at small time lags. Specifically, at certain spatial offsets, the kernel takes negative values, indicating reduced intensity in more distant regions immediately following a seismic event. This pattern qualitatively aligns with stress redistribution effects (so-called stress shadows), which can lead to relative suppression of seismicity in specific areas~\citep{king1994static,harris1998introduction}. Such effects cannot be captured by traditional Hawkes process models that restrict interactions to purely excitatory kernels.
Overall, these findings suggest that our model is capable of uncovering nuanced excitation-inhibition structures and extracting interpretable spatiotemporal interaction patterns from real-world data.

\subsection{Additional Analyses}\label{sect:additional}
Beyond the results above, we provided several additional analyses in the Appendix.
We conducted ablation studies on the quadrature resolution, inducing-grid resolution,
and kernel choice, and found that \ours is robust across these settings
(Appendix~\ref{app:ablation}). We further reported an empirical runtime and memory
analysis as a function of the number of events and grid resolution
(Appendix~\ref{app:runtime}), along with a per-epoch training-time comparison to
neural baselines, where \ours attains competitive efficiency
(Appendix~\ref{app:trainingtime}). To quantify interpretability, we measured
influence-kernel recovery through its correlation with the ground-truth kernel on
synthetic data and a sparsity metric on real data (Appendix~\ref{app:recovery}); 
we additionally verified on a synthetic benchmark with a nonseparable ground-truth
kernel that \ours faithfully recovers nonseparable influence structure despite its
product-kernel covariance (Appendix~\ref{app:nonsep}). 
\section{Conclusion}
We have presented \ours, a Kronecker-structured nonparametric spatiotemporal point process model that enables explicit event-wise relationship discovery while maintaining high modeling flexibility. Empirical evaluations on both synthetic datasets and real-world benchmarks demonstrate promising performance of \ours.

Currently, our method is restricted to rectangular spatial domains due to the tensor-product quadrature construction. In future work, we plan to extend our framework to irregular spatial domains by incorporating triangular finite-element discretizations and adaptive quadrature schemes. Such extensions would enable efficient inference over complex geometries while preserving the numerical stability and computational efficiency. Additionally, exploring sparse-grid quadrature methods may further improve scalability in higher-dimensional settings.

\section*{Acknowledgments}
SZ acknowledges support from NSF CAREER Award IIS-2046295 and NSF CSSI-2311685 (Elements: A Convergent Physics-based and Data-driven Computing Platform for Building Modeling), and NSF DMS-2529112 (Collaborative Research: MATH-DT: Computationally efficient hypercomplex variable-based sensitivity methods for rapid Digital Twin model updating).

\bibliography{ref}

\onecolumn
\appendix
\title{Appendix}
\maketitle
\section{Computational Complexity}\label{sect:complexity}
The time complexity for processing a mini-batch of $B$ sequences, each of length $N$, is $$\Ocal\left(\sum_{k=0}^2 m_k^3 + \sum_{k=1}^2 v_k^3 + BN\left(m(\sum_{i=0}^2 q_i) + v(\sum_{i=1}^2 q_i)\right)\right),$$ where $m=\prod_{k=0}^2m_k$ denotes the total number of mesh points in $\Mcal_f$,  $v=\prod_{k=1}^2v_{k}$ denotes the total number of  mesh points in $\Mcal_g$, and $q_i$ is the number of quadrature nodes along each dimension $i$. The memory complexity is $$\Ocal(\sum\nolimits_{k=0}^2 m_k^2 +\sum\nolimits_{k=1}^2 v_k^2 + m + v + BNq),$$ where $q=\prod_i q_i$ is the total number of nodes in the tensor-product grid $\Qcal$. The dominant memory cost arises from storing the per-dimension kernel matrices and maintaining function evaluations at mesh points, quadrature nodes, and observed events. Overall, the cubic terms depend only on per-dimension mesh sizes rather than the total number of events, enabling scalable mini-batch training and inference.

\section{Dataset Details}\label{sect:dataset}
All the datasets, including the training, validation, and test splits, were downloaded from~\url{https://github.com/tsinghua-fib-lab/Spatio-temporal-Diffusion-Point-Processes/tree/main/dataset}.

\begin{itemize}
    
\item~\textbf{Earthquake}~\citep{chen2021neuralstpp}. 
This dataset contains the time and locations of earthquakes and aftershocks in Japan from 1990 to 2020. Each sequence corresponds to events within one month, with time horizon $T=30$ (time unit: days). The dataset contains 1,050 sequences in total, with sequence lengths ranging from 18 to 543. We used 950 sequences for training, 50 for validation, and 50 for testing.

\item~\textbf{COVID-19}~\citep{chen2021neuralstpp}. 
This dataset records daily COVID-19 cases across counties in New Jersey from March 2020 to July 2020. Each sequence represents events within one week ($T=7$). The dataset contains 1,650 sequences, with sequence lengths varying from 5 to 287. We used 1,450 sequences for training, 100 for validation, and 100 for testing. 
\item~\textbf{Citibike}~\citep{10.1145/3580305.3599511}. This dataset contains bike-sharing records from April to August 2019 in New York City. The time unit is hours, and each sequence represents events within one day ($T=24$). We used 2,440 sequences for training, 300 for validation, and 320 for testing.
\end{itemize}

\begin{figure}[t]
	\centering
	\includegraphics[width=0.5\textwidth]{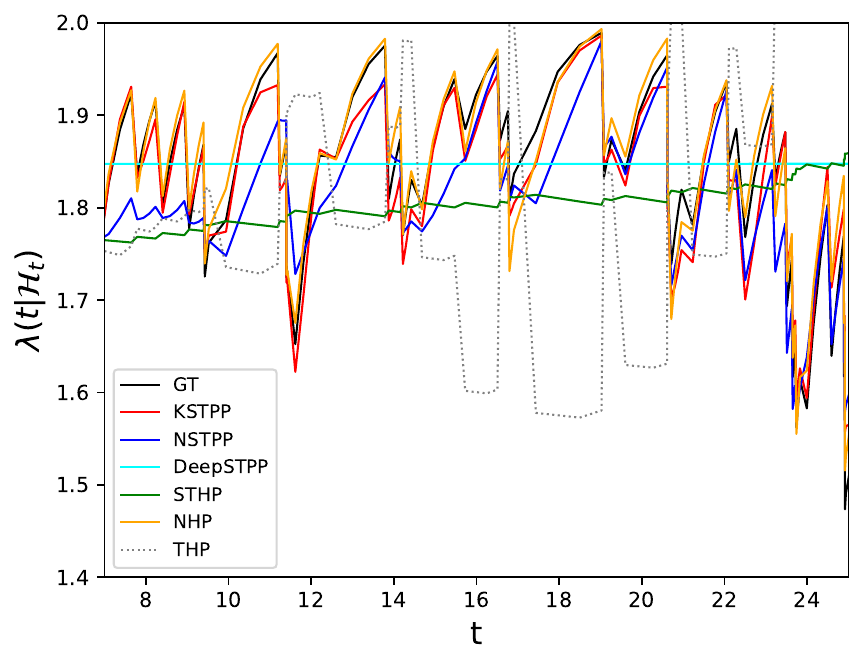}		
	\caption{\small Temporal conditional intensity on example test sequences from \textbf{SYN2}.}\label{fig:lambda-t-syn2}
\end{figure}
\begin{figure*}[!h]
	\includegraphics[width=\textwidth]{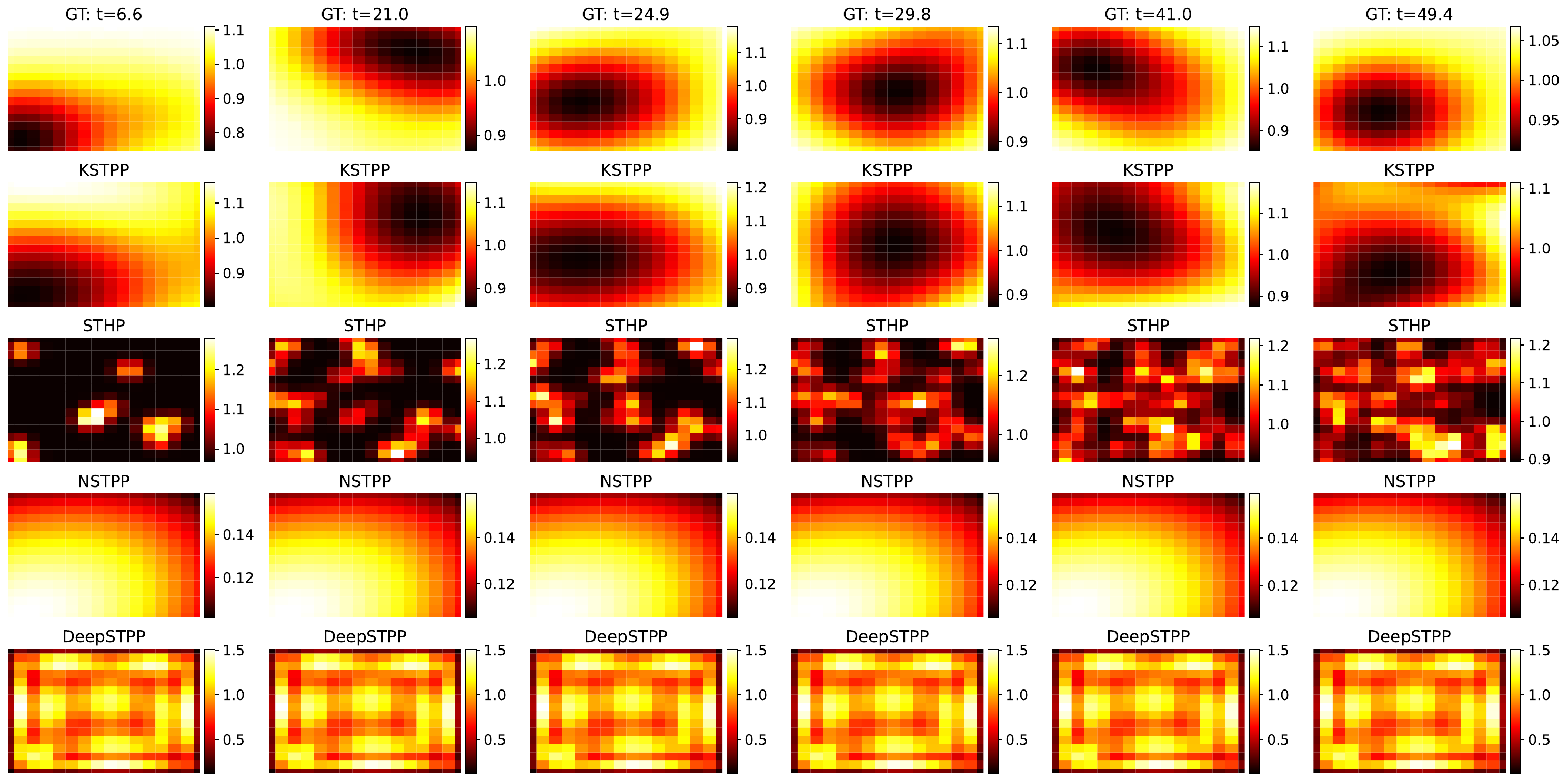}
	\caption{\small Spatial conditional intensities on an example test sequence from \textbf{SYN2} at different time points.}
	\label{fig:spatial-syn2}
\end{figure*}

\begin{figure}[!h]
	\centering
	\includegraphics[width=0.5\columnwidth]{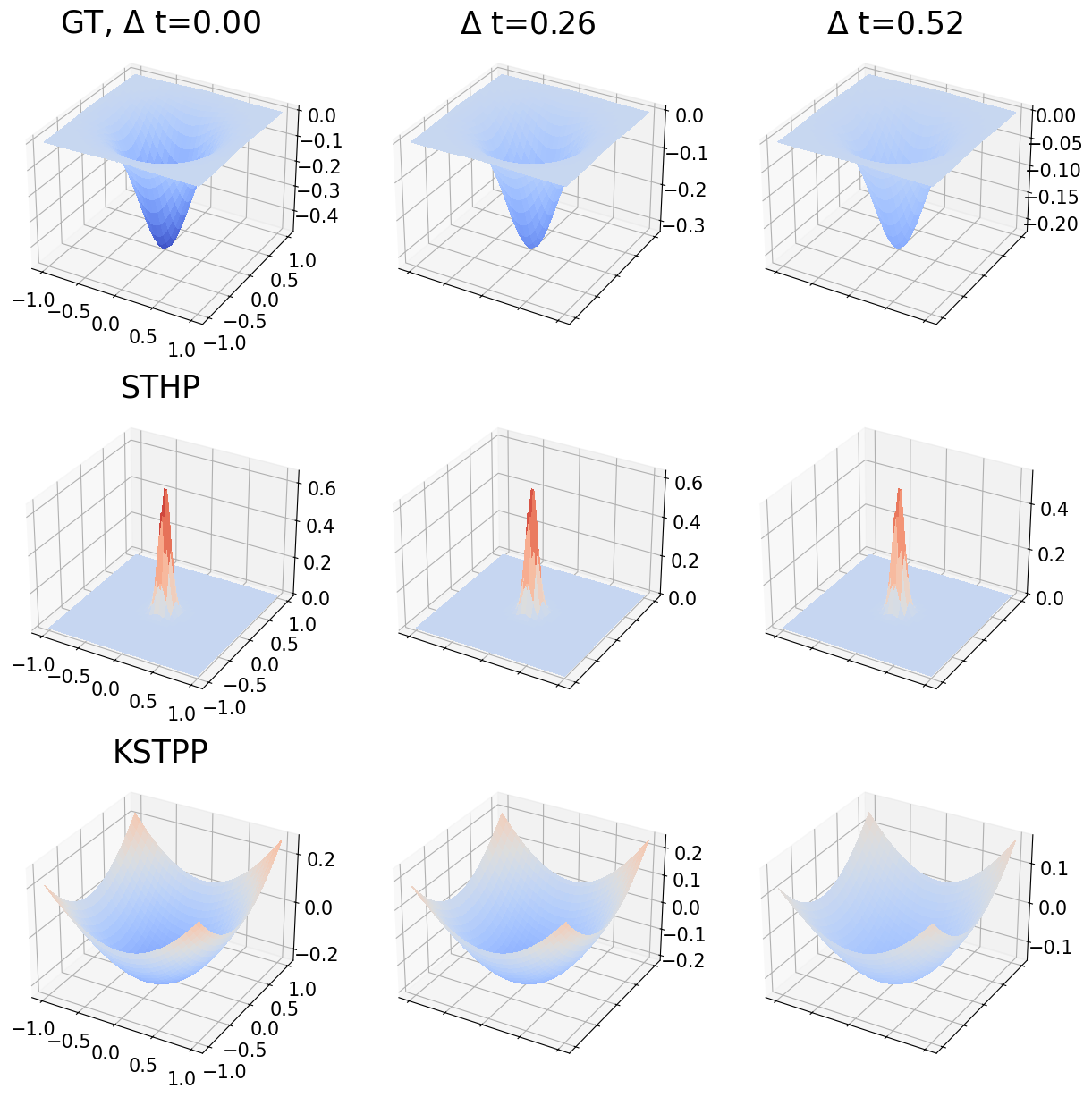}
	\caption{\small Learned influence kernel from \textbf{SYN2} dataset.}
	\label{fig:influence-kernel-syn2}
\end{figure}

\section{Implementation Details and Hyperparameters}
\label{app:impl}
The inducing grid is constructed by taking evenly spaced points along each
input dimension over the corresponding range and forming their Cartesian
product. For all experiments, each kernel is chosen as either the
squared-exponential (SE) kernel or the Mat\'ern-$5/2$ kernel; the kernel
hyperparameters are not manually tuned or selected by cross-validation, but
are instead learned jointly with the remaining model parameters during
training. All models are optimized with Adam using a learning rate of
$10^{-3}$. Table~\ref{tab:hyperparams} summarizes the hyperparameters used by
KSTPP on each real-world dataset. Here $F$ denotes the spatiotemporal inducing
grid used for the influence-kernel GP and $G$ denotes the spatial inducing grid
used for the background-rate GP. We used ReLU activation for the MLP head on the Earthquake dataset, and GELU activation on COVID and Citibike datasets.

\begin{table}[h]
\centering
\begin{tabular}{lllll}
\toprule
Dataset & Kernel & Inducing Grid Size & Quadrature Grid Size & Optimizer / LR \\
\midrule
Earthquake & SE             & $F{:}\,32{\times}32{\times}32$; $G{:}\,12{\times}12$ & $8{\times}12{\times}12$ & Adam / $10^{-3}$ \\
COVID-19   & Mat\'ern-$5/2$ & $F{:}\,32{\times}32{\times}32$; $G{:}\,16{\times}16$ & $8{\times}16{\times}16$ & Adam / $10^{-3}$ \\
Citibike   & Mat\'ern-$5/2$ & $F{:}\,32{\times}32{\times}32$; $G{:}\,12{\times}12$ & $8{\times}12{\times}12$ & Adam / $10^{-3}$ \\
\bottomrule
\end{tabular}
\caption{Hyperparameters used by KSTPP on each real-world dataset.}
\label{tab:hyperparams}
\end{table}

\section{Model Pipeline}
\label{app:pipeline}
\begin{figure}[h]
\centering
\includegraphics[width=\textwidth]{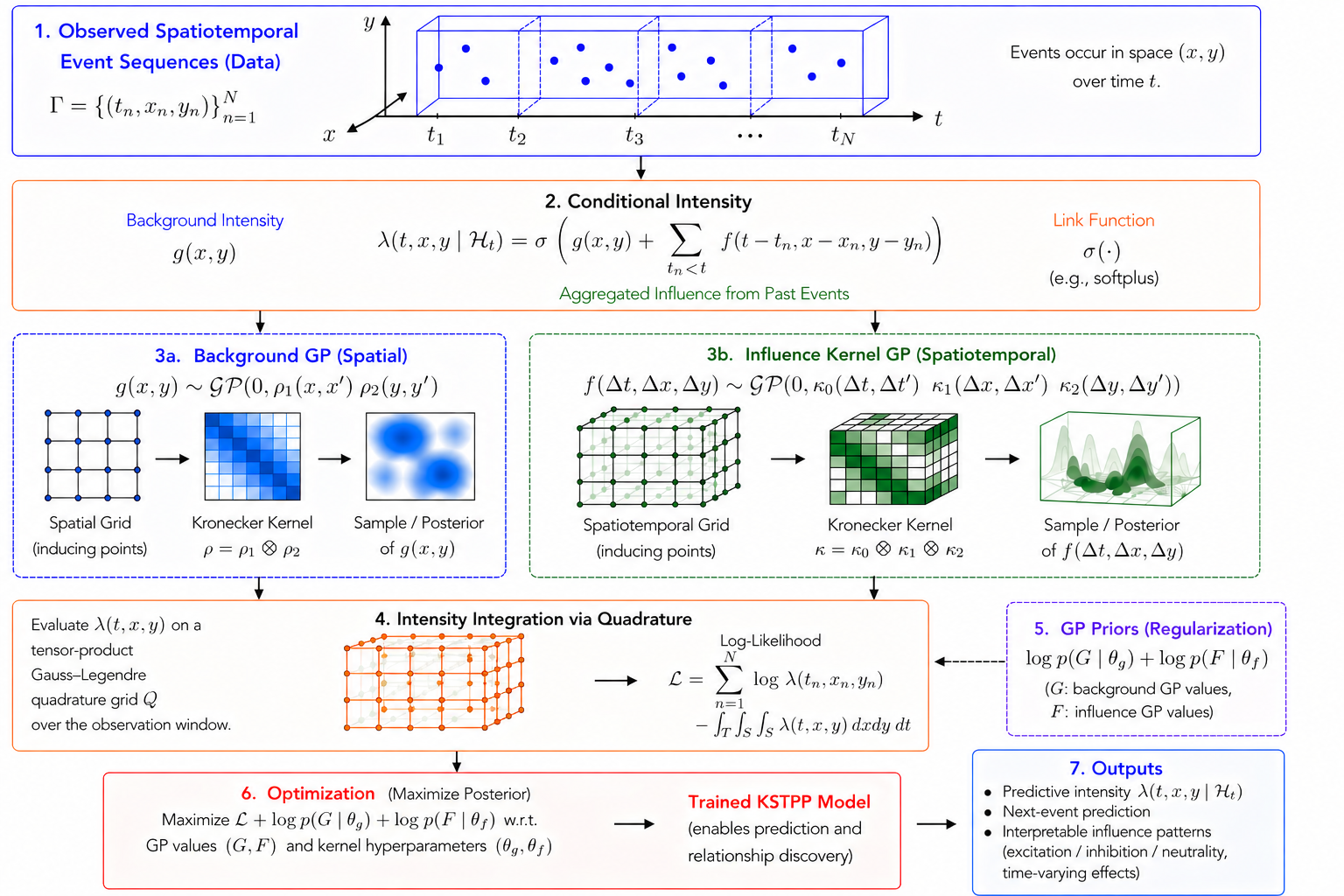}
\caption{Overall KSTPP model pipeline: (1)~observed spatiotemporal event
sequences; (2)~conditional intensity with a background rate, aggregated influence
from past events, and a softplus link function; (3a)~background spatial GP and
(3b)~influence-kernel spatiotemporal GP, both represented on Kronecker-structured
inducing grids; (4)~intensity integration via a tensor-product Gauss--Legendre
quadrature grid for the log-likelihood; (5)~GP priors for regularization;
(6)~optimization of the posterior to obtain the trained KSTPP model; and
(7)~outputs, including predictive intensity, next-event prediction, and
interpretable influence patterns.}
\label{fig:pipeline}
\end{figure}

Figure~\ref{fig:pipeline} illustrates the overall KSTPP pipeline. Starting from
the observed spatiotemporal event sequences, the conditional intensity combines
a background rate with the aggregated influence of past events through a
softplus link function. The background rate and the influence kernel are each
modeled by a Gaussian process whose product kernel induces a Kronecker-structured
covariance on the corresponding inducing grid. The intractable likelihood
integral is approximated with a tensor-product Gauss--Legendre quadrature grid,
and the GP priors act as regularizers. Maximizing the resulting posterior yields
the trained model, which supports predictive intensity estimation, next-event
prediction, and interpretable influence-pattern discovery.

\section{Sensitivity and Ablation Studies}
\label{app:ablation}
We studied the sensitivity of KSTPP to the quadrature resolution, the
inducing-grid resolution, and the kernel choice on the COVID-19 dataset. In each
study, we varied one factor while holding the others fixed. We reported the spatial
prediction error (\emph{Distance}) and the event-time prediction error
(\emph{Time RMSE}); lower is better for both.

\subsection{Quadrature resolution}
\label{app:ablation-quad}
We varied the quadrature grid size while fixing the inducing grids to
$F{:}\,32{\times}32{\times}32$ and $G{:}\,16{\times}16$ and using the
Mat\'ern-$5/2$ kernel. As shown in Table~\ref{tab:abl-quad}, an overly coarse
quadrature grid (e.g., $4{\times}4{\times}4$) yielded a noticeably worse spatial
prediction error, and the spatial performance improved steadily as the
quadrature resolution increased. In contrast, the time RMSE was relatively stable
across resolutions. A likely explanation is that the inter-event time intervals
in this dataset are relatively short, so a moderate number of quadrature nodes is
already sufficient to approximate the temporal contribution to the likelihood.

\begin{table}[h]
\centering
\begin{tabular}{lcc}
\toprule
Quadrature Resolution & Distance & Time RMSE \\
\midrule
$4{\times}4{\times}4$    & 0.445 & 0.103 \\
$8{\times}8{\times}8$    & 0.400 & 0.101 \\
$12{\times}12{\times}12$ & 0.397 & 0.102 \\
$16{\times}16{\times}16$ & 0.393 & 0.102 \\
\bottomrule
\end{tabular}
\caption{Effect of quadrature resolution on prediction error (COVID-19).}
\label{tab:abl-quad}
\end{table}

\subsection{Inducing-grid resolution}
\label{app:ablation-induce}
We varied the inducing-grid resolution while fixing the quadrature grid to
$16{\times}16{\times}16$ and using the Mat\'ern-$5/2$ kernel. The spatial
resolution of $G$ was set equal to that of $F$, so we do not list $G$ separately.
Table~\ref{tab:abl-induce} shows that performance was relatively robust to the
inducing-grid resolution, suggesting that the target latent functions (the
background rate and the influence function) are relatively smooth on this
dataset, so that even a coarse grid such as $4{\times}4{\times}4$ provided
sufficient coverage. Increasing the resolution from $4{\times}4{\times}4$ to
$8{\times}8{\times}8$ gave a slight improvement, while further increasing the
resolution brought no additional gain and could slightly degrade performance,
likely due to the added optimization burden of a larger number of inducing
variables.

\begin{table}[h]
\centering
\begin{tabular}{lcc}
\toprule
Inducing Grid Resolution & Distance & Time RMSE \\
\midrule
$4{\times}4{\times}4$    & 0.393 & 0.0986 \\
$8{\times}8{\times}8$    & 0.391 & 0.0985 \\
$12{\times}12{\times}12$ & 0.391 & 0.0999 \\
$16{\times}16{\times}16$ & 0.394 & 0.101  \\
\bottomrule
\end{tabular}
\caption{Effect of inducing-grid resolution on prediction error (COVID-19).}
\label{tab:abl-induce}
\end{table}

\subsection{Kernel choice}
\label{app:ablation-kernel}
We compared the SE and Mat\'ern-$5/2$ kernels while fixing the inducing grids to
$F{:}\,32{\times}32{\times}32$ and $G{:}\,16{\times}16$ and the quadrature grid
to $8{\times}16{\times}16$ (the first dimension is temporal and the last two are
spatial). As shown in Table~\ref{tab:abl-kernel}, the SE kernel performed slightly
worse than the Mat\'ern-$5/2$ kernel in both spatial and temporal prediction, but
the differences were small. This indicates that both kernels work well on this
dataset and that predictive performance is only mildly sensitive to this choice.

\begin{table}[h]
\centering
\begin{tabular}{lcc}
\toprule
Kernel & Distance & Time RMSE \\
\midrule
Mat\'ern-$5/2$ & 0.392 & 0.100 \\
SE             & 0.393 & 0.101 \\
\bottomrule
\end{tabular}
\caption{Effect of kernel choice on prediction error (COVID-19).}
\label{tab:abl-kernel}
\end{table}

\section{Runtime and Memory Analysis}
\label{app:runtime}
We reported empirical runtime and memory usage as a function of the number of
events and the grid resolution. Runtime is measured in seconds and memory usage
in gigabytes (GB). All experiments were conducted on a Linux workstation with an
NVIDIA H200 GPU. We reported the per-epoch training time, the total prediction
time, and the peak training and test memory.

\subsection{Effect of the number of events}
\label{app:runtime-events}
To isolate the effect of sequence length, we simulated 100 training sequences and
10 test sequences of equal length, varying the length over
$\{50, 100, 150, 200, 250\}$ while fixing both the inducing grid and the
quadrature grid to $8{\times}8$. As shown in Table~\ref{tab:rt-events}, both
runtime and memory grew with the sequence length, because longer sequences
require more likelihood evaluations and more event-history-dependent computation
during training and prediction.

\begin{table}[h]
\centering
\begin{tabular}{lcccc}
\toprule
Seq. Length & Training Time & Prediction Time & Training Memory & Test Memory \\
\midrule
50  & 1.39  & 0.40  & 1.03 & 0.96 \\
100 & 2.43  & 1.18  & 1.72 & 1.44 \\
150 & 4.12  & 3.91  & 3.07 & 2.39 \\
200 & 6.82  & 6.81  & 4.32 & 3.20 \\
250 & 10.37 & 10.61 & 6.37 & 4.61 \\
\bottomrule
\end{tabular}
\caption{Runtime (s) and memory (GB) under different event sequence lengths.}
\label{tab:rt-events}
\end{table}

\subsection{Effect of grid resolution}
\label{app:runtime-grid}
We next varied the grid resolution at a fixed sequence length of 100. First we varied
the quadrature grid size with the inducing grid fixed to $8{\times}8$
(Table~\ref{tab:rt-quad}); then we varied the inducing grid size with the
quadrature grid fixed to $8{\times}8$ (Table~\ref{tab:rt-induce}).

\begin{table}[h]
\centering
\begin{tabular}{lcccc}
\toprule
Quadrature Size & Training Time & Prediction Time & Training Memory & Test Memory \\
\midrule
4  & 1.54 & 1.05 & 1.79 & 1.72 \\
8  & 2.14 & 1.03 & 2.00 & 1.78 \\
12 & 3.20 & 1.20 & 2.27 & 1.74 \\
16 & 3.61 & 1.10 & 2.70 & 1.93 \\
24 & 5.39 & 1.20 & 3.91 & 2.72 \\
\bottomrule
\end{tabular}
\caption{Effect of quadrature resolution on runtime (s) and memory (GB).}
\label{tab:rt-quad}
\end{table}

\begin{table}[h]
\centering
\begin{tabular}{lcccc}
\toprule
Inducing Grid Size & Training Time & Prediction Time & Training Memory & Test Memory \\
\midrule
4  & 0.86 & 0.10 & 1.49 & 1.28 \\
12 & 0.96 & 0.20 & 2.30 & 1.58 \\
16 & 0.95 & 0.20 & 2.50 & 1.84 \\
24 & 0.99 & 0.20 & 3.31 & 1.96 \\
\bottomrule
\end{tabular}
\caption{Effect of inducing-grid resolution on runtime (s) and memory (GB).}
\label{tab:rt-induce}
\end{table}

Increasing either grid resolution raised memory usage and could also increase
runtime, but the runtime growth with grid size was moderate compared with the
growth caused by longer sequences. This is likely because the tensor-product
quadrature computations, the local kernel matrices, and the Kronecker-algebra
operations on the inducing grid parallelize well on the GPU, partially offsetting
the cost of finer grids. In contrast, the number of events had a more pronounced
effect on both runtime and memory, indicating that sequence length is the major
practical driver of computational cost.

\section{Training-Time Comparison with Neural Baselines}
\label{app:trainingtime}
Table~\ref{tab:traintime} reports the per-epoch training time (seconds) of KSTPP
and the neural baselines on the COVID-19 and Earthquake datasets. The per-epoch
training time of KSTPP was comparable to NSTPP and substantially lower than DSTPP,
the diffusion-based approach. DeepSTPP was the fastest, as expected, because it
uses a simpler excitation-only Hawkes-process intensity. Overall, KSTPP attained
competitive training efficiency relative to the neural baselines while
additionally providing flexible and transparent event-wise relationship
discovery.

\begin{table}[h]
\centering
\begin{tabular}{lcc}
\toprule
Method & COVID-19 & Earthquake \\
\midrule
NSTPP    & 21.3 & 25.5 \\
DeepSTPP & 1.3  & 0.8  \\
DSTPP    & 43.9 & 29.3 \\
KSTPP    & 24.5 & 19.7 \\
\bottomrule
\end{tabular}
\caption{Per-epoch training time (seconds).}
\label{tab:traintime}
\end{table}

\section{Quantitative Influence-Kernel Recovery}
\label{app:recovery}

\subsection{Correlation with the ground truth on synthetic data}
\label{app:recovery-corr}
For the synthetic datasets SYN1 and SYN2, we quantified influence-kernel recovery
by the correlation coefficient between the learned influence kernel and the
ground-truth influence kernel, evaluated at several temporal lags $\Delta t$
corresponding to the slices visualized in the main paper. Tables~\ref{tab:corr-syn1}
and~\ref{tab:corr-syn2} report the results for KSTPP and the STHP baseline.

\begin{table}[h]
\centering
\begin{tabular}{lccc}
\toprule
Method & $\Delta t = 0.00$ & $\Delta t = 0.26$ & $\Delta t = 1.77$ \\
\midrule
STHP  & 0.6409 & 0.6409 & $-0.6409$ \\
KSTPP & 0.9932 & 0.9932 & 0.8706 \\
\bottomrule
\end{tabular}
\caption{Correlation between the learned and ground-truth influence kernel on SYN1.}
\label{tab:corr-syn1}
\end{table}

\begin{table}[h]
\centering
\begin{tabular}{lccc}
\toprule
Method & $\Delta t = 0.00$ & $\Delta t = 0.26$ & $\Delta t = 0.52$ \\
\midrule
STHP  & $-0.4379$ & $-0.4379$ & $-0.4379$ \\
KSTPP & 0.7502 & 0.7503 & 0.7507 \\
\bottomrule
\end{tabular}
\caption{Correlation between the learned and ground-truth influence kernel on SYN2.}
\label{tab:corr-syn2}
\end{table}

KSTPP attained substantially higher correlation with the ground truth than STHP,
quantitatively confirming better recovery of the influence function. Moreover,
KSTPP obtained positive correlations across all selected $\Delta t$ values,
indicating stable recovery of the influence-function shape. In contrast, STHP
produced negative correlations on SYN1 at $\Delta t = 1.77$ and on SYN2 at all
selected $\Delta t$ values, indicating that it can recover an incorrect
influence-function shape. SYN2 was more challenging than SYN1 for influence-function
estimation, which was reflected in its lower correlation coefficients.

\subsection{Sparsity structure on real data}
\label{app:recovery-sparsity}
For the real-world Earthquake dataset the ground-truth influence kernel is
unavailable, so we instead reported a sparsity/structure metric for the learned
influence function. Following the $\Delta t$ slices shown in the main paper, we set
a sparsity threshold of $0.05$ and defined the \emph{sparsity ratio} as the
proportion of learned influence-function values whose absolute value falls below
this threshold. Table~\ref{tab:sparsity} reports the sparsity ratio at several
$\Delta t$ values.

\begin{table}[h]
\centering
\begin{tabular}{lcccc}
\toprule
Metric & $\Delta t = 0.00$ & $\Delta t = 0.26$ & $\Delta t = 0.52$ & $\Delta t = 10.35$ \\
\midrule
Sparsity Ratio & 88.6\% & 96.7\% & 99.1\% & 100\% \\
\bottomrule
\end{tabular}
\caption{Sparsity ratio of the learned influence function on the Earthquake dataset.}
\label{tab:sparsity}
\end{table}

The consistently high sparsity ratio indicated that the learned influence function
was highly structured and localized. As $\Delta t$ increased, the sparsity ratio
increased until nearly all function values became very small, showing that the
influence of past events decays over time and becomes negligible after a
sufficiently long interval.

\section{Recovery of a Nonseparable Influence Kernel}
\label{app:nonsep}
The product-kernel covariance used by KSTPP induces a separable covariance
structure, but this does not imply that the represented latent function is
separable. To verify this empirically, we constructed a synthetic dataset whose
influence kernel is explicitly nonseparable. Following the generative form used
for the other synthetic datasets, we set the influence kernel to a zero-mean
Gaussian-shaped function with a dense covariance matrix,
\begin{equation}
f(\Delta t, \Delta x, \Delta y)
= \mathcal{N}\!\big([\Delta t, \Delta x, \Delta y] \mid \mathbf{0}, \Sigma\big),
\qquad
\Sigma =
\begin{bmatrix}
0.08 & 0.04 & 0.02 \\
0.04 & 0.20 & 0.05 \\
0.02 & 0.05 & 0.20
\end{bmatrix}.
\end{equation}
Because $\Sigma$ has nonzero off-diagonal entries, i.e., cross-dimensional
dependencies between the temporal and spatial displacements, this influence
kernel is nonseparable.

\begin{figure}[h]
\centering
\includegraphics[width=\textwidth]{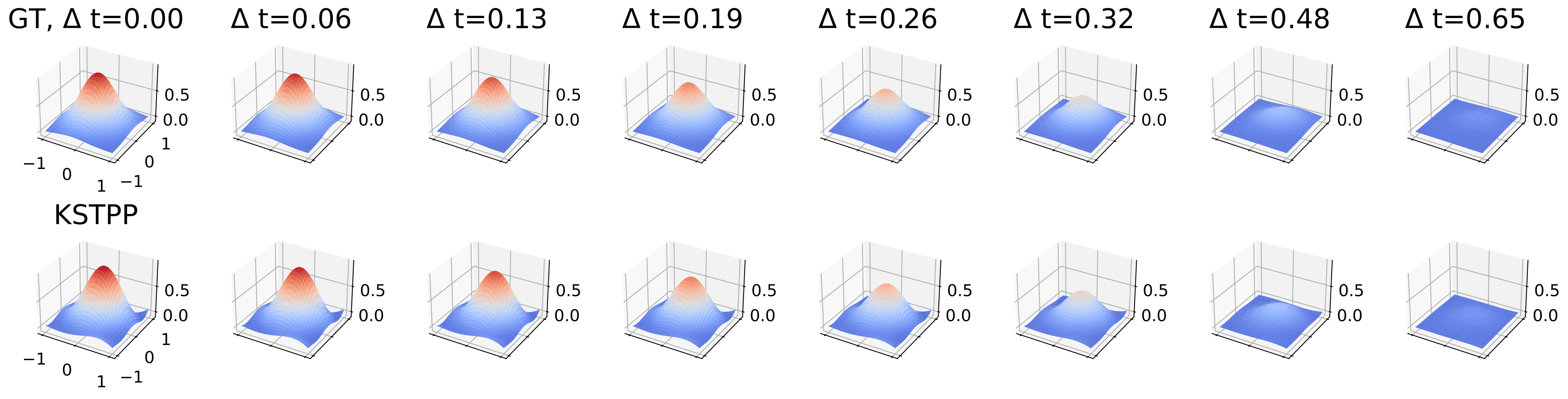}
\caption{Recovery of a nonseparable influence kernel. Top row: the ground-truth
(GT) influence kernel $f(\Delta t, \Delta x, \Delta y)$ shown as spatial surface
slices at increasing temporal lags
$\Delta t \in \{0.00, 0.06, 0.13, 0.19, 0.26, 0.32, 0.48, 0.65\}$. Bottom row: the
corresponding influence kernel recovered by KSTPP at the same $\Delta t$ slices.
KSTPP closely matches the ground-truth shape and its temporal decay despite using
a product-kernel covariance structure.}
\label{fig:nonsep}
\end{figure}

We simulated the same number of training sequences as in SYN1 and SYN2, trained
KSTPP on this dataset, and examined the recovered influence kernel.
Figure~\ref{fig:nonsep} compares the ground-truth influence kernel (top row) with
the kernel recovered by KSTPP (bottom row) across a range of temporal lags
$\Delta t$: KSTPP closely reproduced the ground-truth shape and its temporal
decay. In addition, the relative $L_2$ error for recovering the intensity function
on the held-out test sequences was $0.0361$, which is on the same order as the
relative $L_2$ errors reported for SYN1 and SYN2 in the main paper. These results
provided direct evidence that KSTPP can capture nonseparable influence functions
despite using a product-kernel covariance structure.

\end{document}